\documentclass[preprint,12pt,authoryear]{elsarticle}  

\usepackage{amssymb}
\usepackage{amsthm}

\usepackage{booktabs} 

\usepackage{pifont}

\usepackage{subfigure}  

\usepackage{xcolor}  

\journal{Pattern Recognition}

\begin{document}

\begin{frontmatter}



\title{Learning Data Association for Multi-Object Tracking using Only Coordinates}


\author[poly]{Mehdi Miah}
\author[poly]{Guillaume-Alexandre Bilodeau}
\author[poly]{Nicolas Saunier}

\affiliation[poly]{organization={Polytechnique Montréal},
            addressline={2500 Chemin de Polytechnique},
            city={Montréal},
            postcode={H3T 1J4},
            state={Québec},
            country={Canada}}

\begin{abstract}
We propose a novel Transformer-based module to address the data association problem for multi-object tracking. From detections obtained by a pretrained detector, this module uses only coordinates from bounding boxes to estimate an affinity score between pairs of tracks extracted from two distinct temporal windows. This module, named TWiX, is trained on sets of tracks with the objective of discriminating pairs of tracks coming from the same object from those which are not. Our module does not use the intersection over union measure, nor does it requires any motion priors or any camera motion compensation technique. By inserting TWiX within an online cascade matching pipeline, our tracker C-TWiX achieves state-of-the-art performance on the DanceTrack and KITTIMOT datasets, and gets competitive results on the MOT17 dataset. The code will be made available upon publication.
\end{abstract}

\begin{graphicalabstract}
\includegraphics[trim=0 20 0 0, width=14cm, clip]{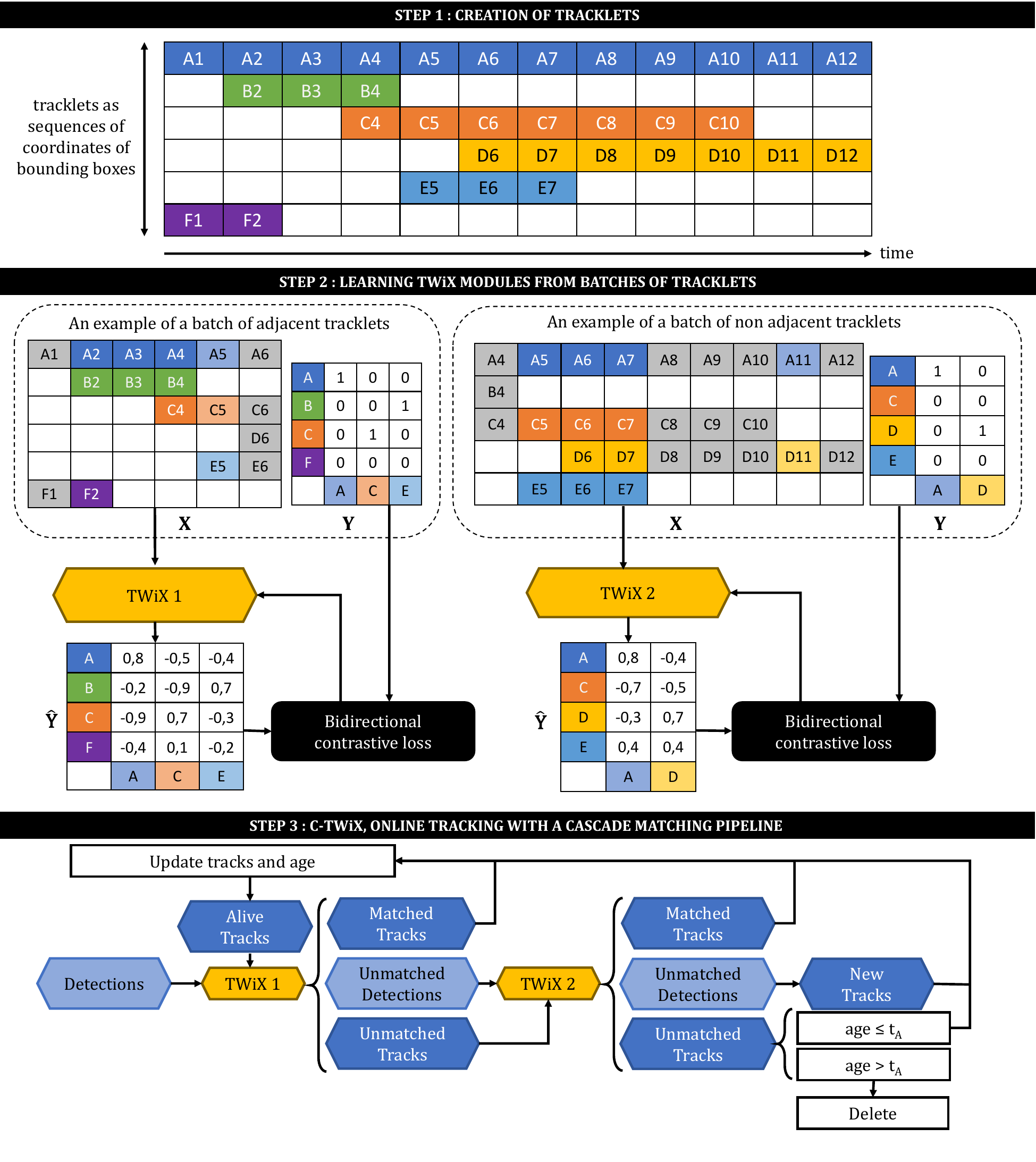}
\end{graphicalabstract}

\begin{highlights}
\item Our Transformer-based model, TWiX, can learn to associate objects using only coordinates;
\item We show that motion priors or intersection-over-union measure are not required for tracking. Using pairs of tracks is sufficient ; 
\item Tracking with TWiX give competitive or state-of-the-results on several dataset.
\end{highlights}

\begin{keyword}
tracking \sep transformer \sep data association \sep motion \sep multi-object tracking



\end{keyword}

\end{frontmatter}



\section{Introduction}
\label{sec:intro}

Multi-object tracking (MOT) consists in detecting all objects of interest, such as cars or pedestrians, and assigning them a unique identity throughout a video. Common applications are road safety analysis~\citep{zangenehpour2016Aresignalized}, video-surveillance and environment awareness in self-driving cars~\citep{sarcinelli2019Handlingpedestrians}. With the improvement of object detectors~\citep{duan2019CenterNetKeypoint,carion2020EndtoEndObject,ge2021YOLOXExceeding}, a popular paradigm to solve MOT is tracking-by-detection, which consists of two steps: detecting objects in each frame of the video and associating detections that correspond to the same object. Under this paradigm, MOT is mainly solved as a data association problem: given two sets of detections, the objective is to find those that refer to the same object. This data association can either be done in an online setting or in an offline setting. In the former one, no information coming from the future can be exploited to track objects. This is the case for real-time applications, like self-driving cars. The offline setting is more suitable for applications such as road traffic and safety analysis. In this setting, it is common to first associate detections between adjacent frames to create tracklets (continuous fragments of trajectories), which are later associated to form complete trajectories of objects.

Usually, given some detections, associations can be made using cues such as appearance (color and texture) and spatio-temporal information (position and motion). Several offline tracking algorithms~\citep{wojke2017Simpleonline,bergmann2019TrackingBells,pang2021QuasiDenseSimilarity} rely on appearance cues, abandoning the motion information. In contrast, many online tracking algorithms~\citep{bewley2016Simpleonline,wang2021TrackAppearance} are only based on motion to make them efficient for real-time applications. 

In the case of occlusions or missed detections, online trackers \emph{generate} new boxes using probabilistic methods or by learning multi-modal distributions of trajectories~\citep{saleh2021ProbabilisticTracklet,tokmakov2021LearningTrack}. This extrapolation may provoke some drifts due to the autoregressive nature of the prediction, or create some false detections. Conversely, it is possible to associate tracklets after an occlusion by solving a simpler \emph{discriminative} task: given a set of tracks and detections, find the ones that correspond to the same object.

These observations raise the following question: is it possible to learn to associate tracklets, without using any appearance information, any camera motion estimation, or any motion prior? In this work, we investigate this question and propose a method to do exactly that. We present TWiX, a \emph{Transformer-based neural network that returns an affinity score between all pairs of tracklets}. It is based on a \emph{supervised contrastive learning task}, where the objective is to discriminate \emph{pairs of tracklets} coming from the same objects from other pairs of tracklets. In TWiX, the representation of a pair of tracklets depends not only on its own characteristics but also on the \emph{context}, represented by the characteristics of other pairs of tracklets. To the best of our knowledge, this is the first work to propose a Transformer-based network to associate tracklets in a discriminative fashion without using any appearance information, motion prior or camera motion compensation technique. 

In summary, we make the following contributions:
\begin{itemize}  
    \item we propose a new association module, named TWiX, that returns a context-dependent affinity score for each pair of tracklets;
    \item we propose a supervised contrastive framework to learn representations for tracklets given their spatio-temporal neighborhood (context);
    \item we obtain state-of-the-art (SOTA) results on DanceTrack and KITTIMOT and competitive results on MOT17, three popular datasets for MOT when using the module TWiX in the cascade matching pipeline;
    \item we conducted several ablation studies and visualizations to show the importance of each component of our method.
\end{itemize}

\section{Related works}

We first introduce some previous tracking-by-detection approaches and their components for data association. Then, we discuss contrastive learning as an extension of re-identification and the use of Transformer networks to solve MOT.

\subsection{Tracking-by-detection and association}

Thanks to improvements in the field of object detection
, tracking-by-detection is a very popular paradigm to solve MOT as an association problem, where the objective is to predict whether two detections belong to the same object. Cues such as appearance and motion are commonly used to compute a similarity score and the Kuhn-Munkres~\citep{kuhn1955Hungarianmethod} algorithm is often used to associate detections in an online fashion. 

First, tracking heavily relies on spatial information to associate objects. SORT~\citep{bewley2016Simpleonline} and ByteTrack~\citep{zhang2021ByteTrackMultiObject} use the Kalman filter~\citep{kalman1960NewApproach} to predict the future position of an object, under a linear constant velocity model, and measure an affinity score with the Intersection over Union (IoU). C-BIoU~\citep{yang2023HardTrack} estimates the future position with a simple linear model and uses a modified version of IoU by proportionally increasing the size of the boxes when computing the affinity score. This allows the model to have a non-zero score even in case where an object moves quickly. IoU-Tracker~\citep{bochinski2017HighSpeedtrackingbydetection} gets rid of any motion model by relying only on IoU to match objects between adjacent frames. However, relying only on spatial positions can provoke some incorrect associations especially in case of occlusions, where objects are naturally close to each other, or when the camera moves.

A strategy to deal with a moving camera is to use a camera motion compensation (CMC) technique~\citep{bergmann2019TrackingBells}. It evaluates the camera motion between two adjacent frames through image registration to adjust the coordinates of bounding boxes. Techniques such as optical flow, enhanced correlation coefficient maximization are required, impacting the speed of the tracking. UCMCTracker proposed a novel camera compensation technique that is applied uniformly throughout each sequence, without computing any frame-by-frame registration~\citep{yi2024UCMCTrackMultiObject}. However, this setting either requires to have access to camera parameters or must be done manually for each sequence.

Another solution is to use visual features in the data association task. The most common methods to extract visual features are based on convolutional neural networks (CNNs) pretrained on a classification task or on a re-identification task~\citep{yu2016POIMultiple,wojke2017Simpleonline}. Such methods extract a visual feature for each detection independently of other detections. However, they also struggle in case of crowded scenes due to total or partial occlusions between objects or in case of different objects with the same appearance~\citep{sun2021DanceTrackMultiObject}.

Conversely, a graph neural network (GNN) is a model where the representation of a node depends on those of its adjacent nodes ~\citep{scarselli2009GraphNeural}. For MOT, GNNs are used to model complex interactions between objects~\citep{braso2020LearningNeural,wang2021JointObject, cetintasUnifyingShortLongTerm2022}. However, GNNs have a limited temporal view span due to their limited scalability. Besides, the interactions between pedestrians can also be learned using a social force model~\citep{liu2021SocialNCE,pellegrini2009Youll}. ~\cite{wang2021DifferentTracking} introduced an attention-based similarity metric that can associate objects by considering an ensemble of detections. Their reconstruction similarity metric is computed by taking into account at once all the current and previous detections.

\subsection{Contrastive learning}

Contrastive learning is an approach to learn representations by making them agree between ``similar instances'' and disagree between ``dissimilar instances''. Re-identification learning with a triplet loss is a special case of contrastive learning where there is one anchor, one positive and one negative instance~\citep{schroff2015FaceNetUnified,hermans2017DefenseTriplet}. By considering multiple negative instances, this loss can be extended to the contrastive loss, as a logistic regression classifier, which learns to discriminate positive pairs from multiple negative pairs~\citep{gutmann2010Noisecontrastiveestimation,mnih2013Learningword}. This contrastive loss is widely used in the field of unsupervised representation learning~\citep{vandenoord2018RepresentationLearning,chen2020SimpleFramework,grill2020BootstrapYour} and recently for online MOT to learn the representation of a track~\citep{li2021SemiTCLSemiSupervised}. Some representations are also trained with multiple positive pairs~\citep{frosst2019AnalyzingImproving,pang2021QuasiDenseSimilarity}. 
Data augmentation, large batch size and hard negative mining (a technique used to focus the learning on the most difficult examples) are commonly listed as critical components in contrastive learning to extract good representations~\citep{appalaraju2020GoodPractices}. In computer vision, such augmentations are for example image cropping, color dropping, color distortion and random Gaussian blur~\citep{chen2020SimpleFramework}. A large batch size ensures that the contrastive loss encounters enough negative samples, and the hard negative mining helps to learn more discriminative representations~\citep{grill2020BootstrapYour}. 

Intuitively, a contrastive framework aims at finding positive pairs among a large collection of negative pairs. In MOT, GNNs offer a natural way to create a few positive pairs (detections from the same object) and many negative pairs (detections from two different objects). GNNs were used to learn representations for all detections of a given object based on all detections from its previous and next frames~\citep{braso2020LearningNeural}. Our framework differs from this previous work by learning representations at the level of a pair of tracklets instead of the detection level, and by formulating the problem as a contrastive task instead of a classification task. 




\subsection{Transformers for tracking}

Transformer networks are based on an attention mechanism to extract relevant representations from sequences~\citep{vaswani2017AttentionAll}. Even if this network was originally proposed for natural language processing in which sentences are the input sequences~\citep{devlin2019BERTPretraining}, a Transformer network can be exploited for image classification~\citep{dosovitskiy2021ImageWorth} or object detection~\citep{carion2020EndtoEndObject} if an image is reshaped as a sequence of patches. The first use of a Transformer for solving MOT is the work of TransTrack~\citep{sun2021TransTrackMultiple}. It jointly detects and associates objects in an online fashion by using object features from previous frames as queries of the current frame. Similarly, Trackformer was proposed to solve multi-object tracking and segmentation~\citep{meinhardt2021TrackFormerMultiObject}. This online algorithm introduced the track queries to refer to an object in a video. MOTR used a Transformer decoder to model a full track in an end-to-end manner~\citep{zeng2021MOTREndtoEnd}. It showed good ability to associate without using any re-identification method or track non-maximum suppression.

Our work distinguishes itself from these previous works by considering the association step as a \emph{pure discriminative task, without generating any positions (for example with a Kalman filter or linear motion model), without using any appearance information and without using the IoU at the inference step}.

\section{TWiX}

We propose TWiX, a Transformer-based module that returns an affinity matrix between two sets of tracklets selected from two temporal windows. It first considers the interaction between observations within two tracklets, then the interaction between all the pairs of tracklets. TWiX is based on contrastive learning to extract representations taking into account the spatio-temporal coordinates of all the tracklets. For that, given some detections, we first create tracklets. Then using two temporal windows, each containing a set of tracklets, we create all pairs of tracklets from the two sets and we use a contrastive loss to learn whether a given pair of tracklets comes from the same object given all other pairs of tracklets to complete the multiple object tracking process. 

\subsection{Creation of tracklets}
\label{sec:creationTracklets}

The tracklets are created from detected bounding boxes and \emph{not} from any ground-truth positions. Indeed, heavy data augmentations are a key aspect in the use of a contrastive framework. Detections are a natural way to augment the data: bounding box coordinates are noisier, some boxes are missing (false negative), others are extra (false positive).

Given a video and a set of detections, we create tracklets by associating bounding boxes between adjacent frames. Many approaches are possible like associating detections from frame $t$ and $t+1$ that share the highest IoU or a buffered version of IoU~\citep{yang2023HardTrack}. This IoU can also be computed after warping the detections with the optical flow. The Kalman filter can also be used to leverage object motion. We use the IoU and Kuhn-Munkres algorithm on the cost matrix made of negative IoUs. During this short-term association between adjacent frames, an association is kept as long as the IoU is higher than a threshold $\theta_s$.

\subsection{Creation of a batch of tracklets}

Once tracklets are created, as illustrated in Figure~\ref{fig:creationData}, we define a batch of tracklets as follows, without loss of generality:

\begin{figure}[ht]
    \centering
    \subfigure[]{\includegraphics[scale=0.5, trim=0 270 500 0, clip]{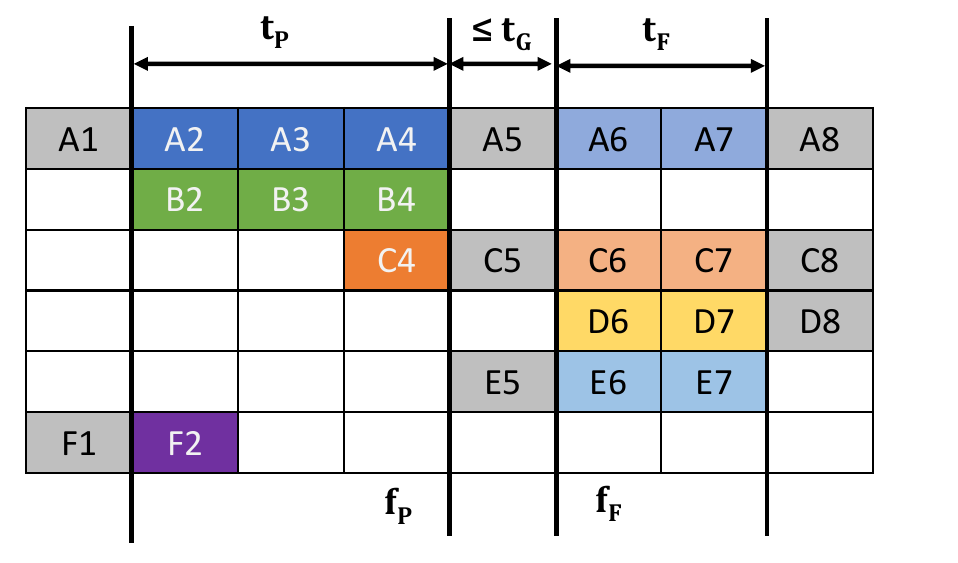}}
    \subfigure[]{\includegraphics[scale=0.5, trim=0 330 780 0, clip]{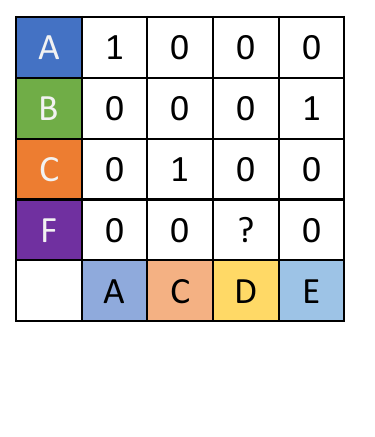}}
    \caption{Creation of a batch of tracklets with two temporal windows used during the training. The two frames of reference are $f_P$=4 and $f_F$=6 and the temporal windows are of length $t_P$=3 and $t_F$=2 frames. a) The set of past and future tracklets contains each four tracklets. Gray detections are completely ignored in this batch of tracklets. b) The matrix $\textbf{Y}$ indicates whether a pair is positive (1), negative (0) or ignored (?). Best viewed in color.}
    \label{fig:creationData}
\end{figure}

\begin{itemize}
    \item two frames, $f_P$ (reference frame of the past) and $f_F$ (reference frame of the future), are selected such that $t_G\times fps \geq f_F - f_P - 1 \geq 0 $, where $t_G$ is the maximal temporal gap between tracklets and $fps$ the number of frames per second in the video;
    \item sub-tracklets of the video that appears between the frames $f_P - t_P \times fps $ and $f_P$ form the set of past tracklets $\mathcal{S}_P$, and those between the frames $f_F$ and $f_F + t_F \times fps$ form the set of future tracklets $\mathcal{S}_F$, where $t_P$ (resp. $t_F$) is the maximal temporal duration on tracklets of the past (resp. future);
    \item from sets $\mathcal{S}_P$ and $\mathcal{S}_F$ containing respectively $n_P$ and $n_F$ tracklets, we create a matrix $\textbf{Y} \in \mathbb{R}^{n_P \times n_F}$ that indicates if a pair of sub-tracklets $(T^i, T^j) \in \mathcal{S}_p \times \mathcal{S}_f$ corresponds to the same object. 
\end{itemize}

Any pair of sub-tracklets is either positive, negative or ignored. A positive pair is obtained if the two sub-tracklets belong to the same ground-truth object, or are extracted from the same tracklet. A negative pair is obtained if the two sub-tracklets do not belong to the same ground-truth object or their full tracklets have some temporal overlap. The remaining pairs are labeled as ignored, for instance if one tracklet falls into some ignored regions or when the two sub-tracklets do not match with any ground-truth object.


\subsection{Neural Network Architecture of TWiX}

\begin{figure*}[ht] \centering
    \includegraphics[trim=80 00 0 0, width=14cm, clip]{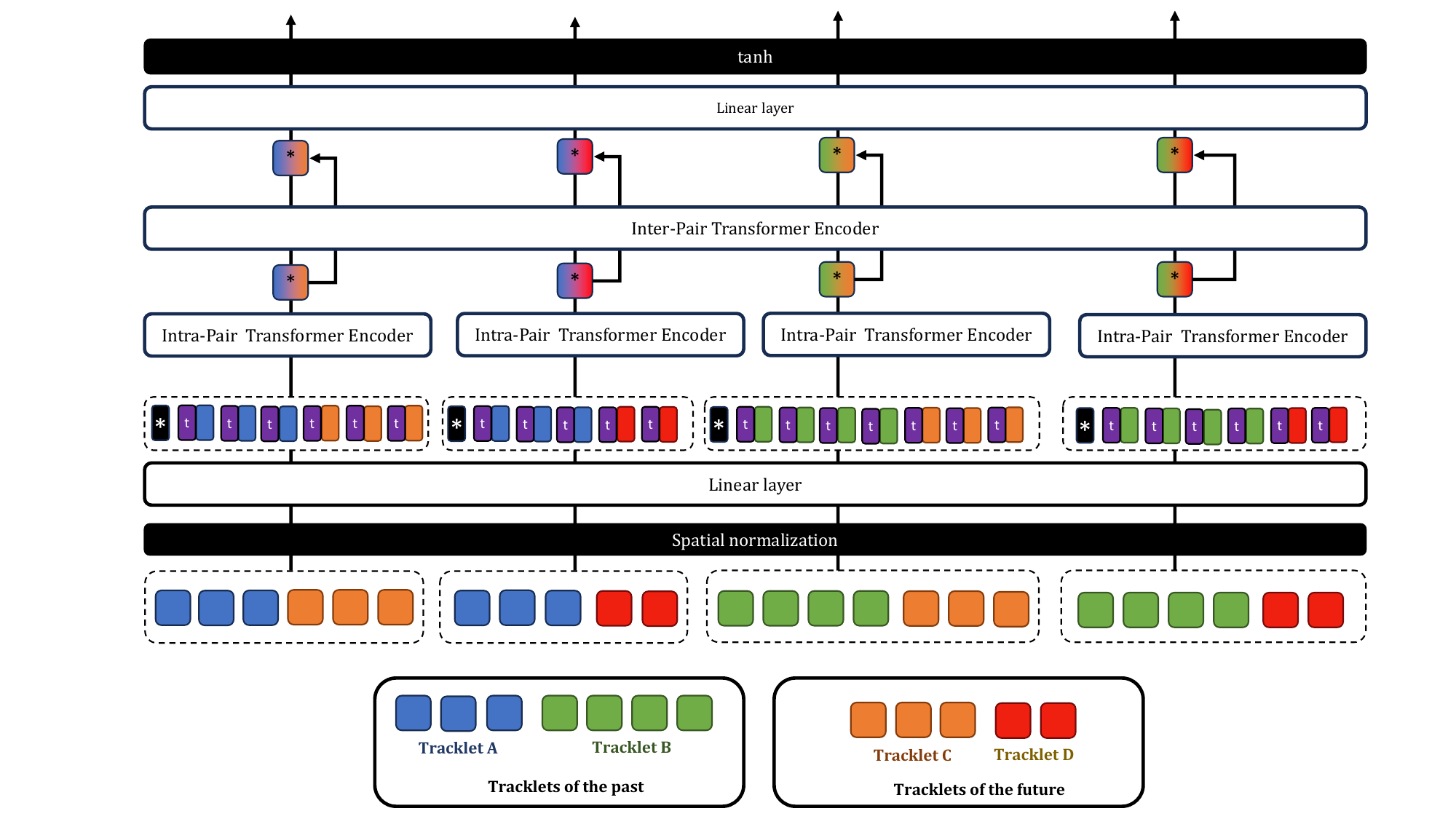}
    \caption{Architecture of TWiX (read from bottom to top). First, pairs of tracklets are normalized and linearly projected then encoded with a Transformer where attention is applied on the temporal dimension. Then, refined representations are obtained with a second Transformer which pays attention to all other pairs. Finally, a linear layer and a hyperbolic tangent function are used to compute an affinity score for each pair. Best viewed in color.}
    \label{fig:model}
\end{figure*}

Figure~\ref{fig:model} illustrates the TWiX model. The inputs of TWiX are two sets of tracklets: those of the past and those of the future. A tracklet $T$ of length $W$ is described by two sequences: a matrix of coordinates $\mathbf{C} \in \mathbb{R}^{W \times 4}$ and a vector of timestamps $\mathbf{T} \in \mathbb{R}^{W}$. We create \emph{pairs} of tracklets by combining each tracklet from the past with those from the future. The motivation of using pairs of tracklets instead of tracklets is to get more discriminative features. Indeed, in the case of an occlusion, coordinates of bounding boxes of involved objects are very similar, so the representations of each tracklet may also be very similar if considered \emph{independently}. By considering all pairs of tracklets, it is possible to get more discriminative representations for each pair by considering the \emph{whole context}. We note that considering pairs is similar that what is done when computing IoU, which is very effective to associate detections.

After concatenating coordinates for each a pair, they are normalized following a minmax scaler such that they are strictly between -1 and 1. We decided to use a minmax normalization instead of a normalization based on the full size of the image to generate more diversified training data where coordinates are changing depending on the position of other objects. These spatial coordinates are then linearly mapped to a $D$-dimensional vector, where $D$ is the dimension of the representation. 

Within the pair, temporal information is added using a fixed positional encoding based on sine and cosine functions~\citep{vaswani2017AttentionAll}. This temporal information is the temporal distance of each observation with the first observation of the second track. Then, a \texttt{[CLS]} token is concatenated for each pair of tracklets~\citep{devlin2019BERTPretraining,dosovitskiy2021ImageWorth}. A Transformer Encoder, named Intra-Pair, extracts a representation for each pair where the attention spans on the temporal dimension. Finally, a second Transformer Encoder, named Inter-Pair, extracts a representation for each pair taking into account the representations of all other pairs. These second representations are added to the first ones using a skip connection. At this step, we have a tensor of shape $n_P \times n_F \times D$. To have an affinity matrix, this tensor is sent to a linear layer and a hyperbolic tangent activation function to obtain a matrix of shape $n_P \times n_F$ with values between -1 and 1. 

We note that for contrastive learning, the loss function is usually computed using the cosine similarity of embeddings~\citep{chen2020SimpleFramework}. Here, since the pairs are computed inside the model and not during the back-propagation step, we use a hyperbolic tangent function to obtain affinities between -1 and 1.

\subsection{Contrastive loss}
\label{sec:loss}

The output of TWiX is an affinity matrix between all tracklets of the past and of the future. The weights of TWiX, are learned by minimizing a bidirectional contrastive loss, $\mathcal{L}_{bdrC}$, defined as the sum of the forward, $\mathcal{L}_{fwdC}$, and the backward contrastive loss, $\mathcal{L}_{bwdC}$.


Formally, the contrastive loss function $l_C$ for a single positive pair with score $s^+$ relatively to a set of negative pairs $\mathcal{N}$ of size $N_{neg}$ is defined as follows,

\begin{equation}
l_C(s^+, \mathcal{N}) = \log \left( 1 + \frac{B}{N_{neg}}\sum \limits_{s^- \in \mathcal{N}} exp \left[-\frac{s^+-s^-}{\tau }\right] \right),
\label{eq:loss}
\end{equation}

where $\tau$ denotes a temperature parameter and $B$ a batch size parameter.


If $\hat{\textbf{Y}} = (\hat{y}_{ij})_{1 \leq i \leq n_P \atop 1 \leq j \leq n_F}$ and $\textbf{Y} = (y_{ij})_{1 \leq i \leq n_P \atop 1 \leq j \leq n_F}$ are the predicted affinity matrix and the ground truth matrix, $N^+$ the number of positive pairs in $\textbf{Y}$, the forward and backward contrastive losses are defined as follows,

\begin{equation}
\mathcal{L}_{fwdC}(\hat{\textbf{Y}}, \textbf{Y}) = \frac{1}{N^+} \sum \limits_{i, j \atop y_{ij} = 1} l_C(\hat{y}_{ij}, \{ \hat{y}_{il} | 1 \leq l \leq n_F, y_{il} = 0\}), 
\label{eq:fwdC}
\end{equation}

\begin{equation}
\mathcal{L}_{bwdC}(\hat{\textbf{Y}}, \textbf{Y}) = \frac{1}{N^+} \sum \limits_{i, j \atop y_{ij} = 1} l_C(\hat{y}_{ij}, \{ \hat{y}_{kj} | 1 \leq k \leq n_P, y_{kj} = 0\}), 
\label{eq:bwdC}
\end{equation}

The bidirectional loss is finally defined as, 
\begin{equation}
\mathcal{L}_{bdrC} = \mathcal{L}_{fwdC} + \mathcal{L}_{bwdC}.
\label{eq:loss_bidirectional}
\end{equation}





In contrast to typical contrastive frameworks used for visual pretraining~\citep{chen2020SimpleFramework}, here the number of tracklets inside the batch is not fixed, hence the shape of the ground truth matrix is also not fixed. So we scale the number of negative pairs with the batch size parameter $B$ ($B \gg N_{neg} $) to simulate a large batch size. 




The intuition behind the contrastive loss is to facilitate the learning of discriminative features. This ensures that the affinity between two tracklets referring to the same object is higher than the affinities between the first tracklet and all other tracklets. Introducing bidirectionality increases the robustness of the calculated affinities by incorporating both tracking and reverse tracking aspects.

\subsection{Tracking with TWiX}

A trained TWiX module outputs an affinity matrix between two sets of tracks. This module can replace any module to compute similarities, such as IoU measure, in any tracking pipeline. Hence, our TWiX module can create tracklets without using IoU.

\begin{figure*}[ht] \centering
    \includegraphics[trim=0 790 0 0, width=14cm, clip]{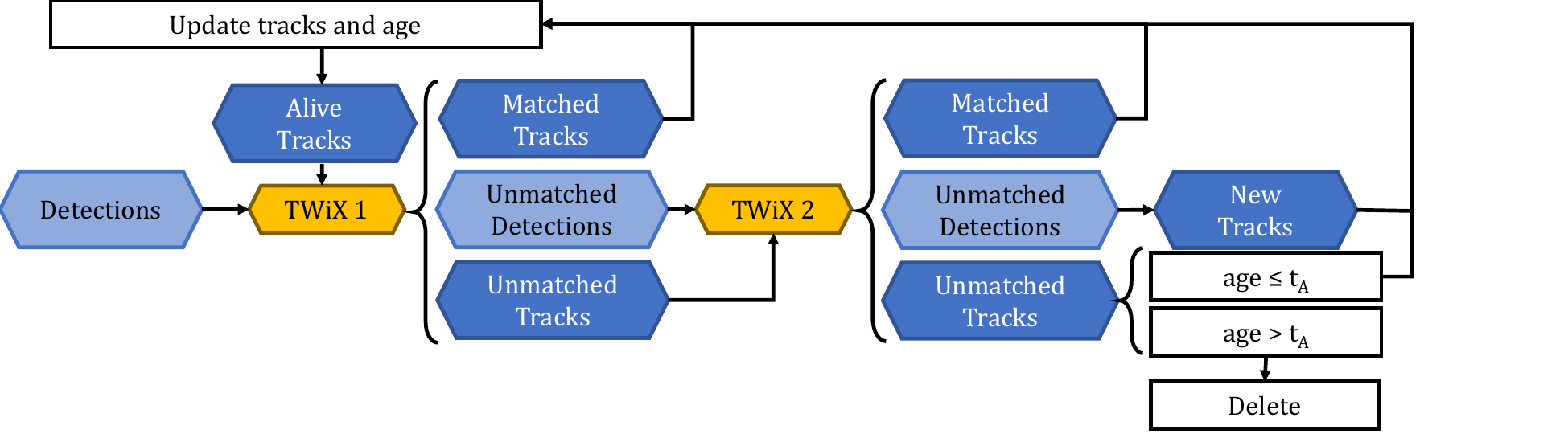}
    \caption{Our tracker C-TWiX use a cascade matching pipeline for tracking. The BIoU-computed matrix in C-BIoU is replaced by our TWiX module. Best viewed in color.}
    \label{fig:pipeline}
\end{figure*}

For online tracking, we use a MOT pipeline similar to cascade matching as described in C-BIoU~\cite{yang2023HardTrack} by only replacing the BIoU-computed matrix by the one obtained with the TWiX module. In this pipeline, as illustrated in Figure~\ref{fig:pipeline}, the first TWiX module matches alive tracks (as the set of past tracklets $\mathcal{S}_P$) with detections (as the set of future tracklets $\mathcal{S}_F$) , then the second TWiX module matches unmatched tracks (as the set of past tracklets $\mathcal{S}_P$) with unmatched detections (as the set of future tracklets $\mathcal{S}_F$). The Hungarian algorithm is used for both matching, if the score is higher than a threshold $\theta_1$ for the first match and $\theta_2$ for the second one. We create new tracks from unmatched detections whose score are higher than $\theta_T$. Tracks are killed when their age is higher than $t_A$.

We note that the TWIX module can also be applied for offline tracking. In that case, we need to adapt the association pipeline by first associating detections between adjacent frames to create tracklets. Then a second matching would associate tracklets as long as their affinity is high. 

As the SOTA methods considered for comparison were developed only for online tracking, in order to have a fair comparison with them, we only applied the online tracking pipeline in the following experiments.

\section{Experiments}
We conducted experiments on three datasets focused on pedestrians and cars. MOT17 is a popular dataset with 7 training videos and 7 test videos ~\citep{milan2016MOT16Benchmark}. They contain scenes with simple and linear movements of humans. The camera is either fixed or mounted on a car or carried by a pedestrian. DanceTrack dataset was recently proposed to be more challenging: targets have similar appearance, are subject to severe occlusions and have irregular motions ~\citep{sun2021DanceTrackMultiObject}. It contains 40 videos for training, 25 for validation and 35 for test. Finally KITTIMOT contains videos of cars and pedestrians~\citep{geiger2012Arewe}. A camera is mounted on top of a car, driving through a city. This dataset is challenging because the videos are recorded at 10 FPS, whereas it is 20 for DanceTrack and between 14 and 30 for MOT17.

We follow the common practices in multi-object tracking on MOT17 and KITTIMOT where official validation sets are not provided. We split the full training sequences of MOT17 into two halves for training and for validation following~\cite{zhou2020TrackingObjects}. And we use the splits from KITTIMOTS, the pixel-level variant of KITTIMOT, to create the training and the validation sets of KITTIMOT, following~\cite{luiten2020HOTAHigher}. 

The quality of a tracker is measured on its ability to detect objects of interest and on its ability to keep the identity consistent throughout the video. We adopted the higher order metric for multi-object tracking (HOTA) to evaluate the quality of the tracking~\citep{luiten2020HOTAHigher}. This metric was introduced to complement the MOTA~\citep{stiefelhagen2007CLEAR2006} metric that takes into account the quality of the detections over the quality of association. The HOTA metric decomposes into the detection accuracy (DetA) and the association accuracy (AssA). Nevertheless, we provide also the MOTA and IDF1~\citep{ristani2016PerformanceMeasures} as additional metrics.

\subsection{Implementation details}

For a fair comparison, we used the detections from YOLOX~\citep{ge2021YOLOXExceeding} with weights provided by ByteTrack on the datasets DanceTrack and MOT17. Following~\cite{caoObservationCentricSORTRethinking2022}, we use the detections from PermaTrack~\citep{tokmakov2022ObjectPermanence} on the KITTIMOT dataset. For all datasets, only detections with a score higher than 50\% and bigger than 128 pixels are kept. 

On MOT17 and KITTIMOT, after the selection of hyper-parameters on the validation set, we re-trained the two modules on the full training sets.

TWiX extracts representations for pairs of tracks taking into account their relative surroundings. However, these surroundings depend on the matching step. In the pipeline based on cascade matching, we noticed that associations at the first matching occurred mainly between adjacent tracklets, when the temporal gap is zero. That is why we train the TWiX module with batches of tracklets with $t_G = 0$ for the first matching step and with a non-zero $t_G$ for the second step. We note that to train the TWiX module for the second matching, we only consider batches where $f_P$ (resp. $f_F$) corresponds to an end (resp. a beginning) of a tracklet. Indeed, during a partial occlusion, spatial coordinates experience a disturbance because the size of a bounding box decreases. Since no appearance information is used, such a signal is important to learn how to associate tracklets under occlusion. Other hyper-parameters are described in Table~\ref{tab:param} for each dataset. Please note that only one frame is used for the tracklets of the future, as they are detections.

\begin{table}[ht]
\begin{center}
\caption{Hyper-parameters for each dataset used during the training and inference steps.}
\label{tab:param}
\begin{tabular}{lccc}
Parameter & DanceTrack & MOT17 & KITTIMOT  \\
\toprule
Maximal temporal gap $t_G$ &  1.6 sec  & 0.8 sec  & 0.8 sec \\
Past window size $t_P$ &  0.8 sec  & 0.4 sec  & 0.4 sec \\
Future window size $t_F$ &  $\frac{1}{fps}$ sec  & $\frac{1}{fps}$ sec  & $\frac{1}{fps}$ sec \\
Matching thresholds $(\theta_1, \theta_2)$ & $(-0.5, -0.2)$ &  $(0.9, -0.5)$ &  $(0.4, -0.6)$\\
Maximum age $t_A$ &  1.6 sec  & 0.8 sec  & 0.8 sec \\
Minimal score $\theta_T$ &  90\%  & 70\%  & 50\% \\
\bottomrule
\end{tabular}
\end{center}
\end{table}

As for the TWiX architecture, the hyperparameters are the same for the two Transformer encoders: the dimension size is 32, the number of heads is 16 and the dimension of the feedforward layer is 32. The batch size and temperature in the loss are respectively $B=1024$ and $\tau=0.1$. The model is trained during 30 epochs with the Adam optimizer~\citep{kingma2015AdamMethod}. The only difference between the first and the second TWiX modules is that the first one is trained with single-layer Transformers and a learning rate of $1e-4$, and the second with Transformers with four layers and a learning rate of $1e-3$.

\subsection{Main results}

Tables~\ref{tab:results_test_dt}, \ref{tab:results_test_mot17} and \ref{tab:results_test_kittimot} contain respectively the performance measures of our tracker C-TWiX (TWiX in a cascade matching algorithm) on DanceTrack, MOT17 and KITTIMOT datasets. By analyzing the results, we can first notice that our tracker C-TWiX outperforms other appearance-free trackers on DanceTrack. This dataset is particularly challenging due to the irregular motions of persons. Previous methods employ linear filters, such as a Kalman filter or a linear model, to estimate the new position. Our tracker does not use such assumptions, which are not always correct, enabling it to improve the association step, increasing the AssA score by 1.8 points and the HOTA by 1.5 points. Only UCMCTracker has a better HOTA. However, this tracker requires to manually select the compensation parameters on each sequence individually. Our tracker C-TWiX does not require such sequence-level adjustment.

\begin{table}[ht]
\begin{center}
\caption{Performance on the test set of DanceTrack. Only trackers using the detections from ByteTrack and using only coordinates are shown. \textbf{\textcolor{red}{Bold red}} and \textit{\textcolor{blue}{italic blue}} indicate respectively the first and second best methods within each category.}
\label{tab:results_test_dt}
\begin{tabular}{lp{0.8cm}p{0.8cm}p{0.8cm}p{0.8cm}p{0.8cm}}
Method & \footnotesize{HOTA} & \footnotesize{DetA} & \footnotesize{AssA} & \footnotesize{MOTA} & \footnotesize{IDF1}  \\
\midrule
\multicolumn{6}{l}{\tiny{METHODS USING MOTION COMPENSATION}} \\
SparseTrack~\citep{liu2023SparseTrackMultiObject} & \textit{\textcolor{blue}{55.5}} & \textbf{\textcolor{red}{78.9}} & \textit{\textcolor{blue}{39.1}} & \textbf{\textcolor{red}{91.3}} & \textit{\textcolor{blue}{58.3}} \\
UCMCTrack~\citep{yi2024UCMCTrackMultiObject} & \textbf{\textcolor{red}{63.4}} & NA & \textbf{\textcolor{red}{51.1}} & \textit{\textcolor{blue}{88.8}} & \textbf{\textcolor{red}{65.0}} \\
\midrule
\multicolumn{6}{l}{\tiny{METHODS NOT USING ANY MOTION COMPENSATION}} \\
SORT~\citep{bewley2016Simpleonline}             & 50.0 & 75.5 & 33.2 & 90.4 & 52.0  \\
DeepSORT~\citep{wojke2017Simpleonline}         & 45.6 & 71.0 & 29.7 & 87.8 & 47.9  \\
ByteTrack~\citep{zhang2021ByteTrackMultiObject}        & 51.9 & 80.1 & 33.8 & 90.9 & 52.0 \\
OC-SORT~\citep{caoObservationCentricSORTRethinking2022}          & 54.6 & 80.4 & 40.2 & 89.6 & 54.6 \\
MotionTrack~\citep{xiao2023MotionTrackLearning} & 52.9 & 80.9 & 34.7 & 91.3 & 53.8 \\
C-BIoU~\citep{yang2023HardTrack}           & \textit{\textcolor{blue}{60.6}} & \textit{\textcolor{blue}{81.3}} & \textit{\textcolor{blue}{45.4}} & \textbf{\textcolor{red}{91.6}} & \textit{\textcolor{blue}{61.6}}  \\
C-TWiX (ours)           & \textbf{\textcolor{red}{62.1}} &\textbf{\textcolor{red}{81.8}} & \textbf{\textcolor{red}{47.2}} & \textit{\textcolor{blue}{91.4}} & \textbf{\textcolor{red}{63.6}} \\
\bottomrule
\end{tabular}
\end{center}
\end{table}

\begin{table}[ht]
\begin{center}
\caption{Performance on the test set of MOT17. Only trackers using the detections from ByteTrack and using only coordinates are shown. \textbf{\textcolor{red}{Bold red}} and \textit{\textcolor{blue}{italic blue}} indicate respectively the first and second best methods within each category.}
\label{tab:results_test_mot17}
\begin{tabular}{lp{0.8cm}p{0.8cm}p{0.8cm}p{0.8cm}p{0.8cm}}
Method &  \footnotesize{HOTA} & \footnotesize{DetA} & \footnotesize{AssA} & \footnotesize{MOTA} & \footnotesize{IDF1}  \\
\midrule
\multicolumn{6}{l}{\tiny{METHODS USING MOTION COMPENSATION}} \\
BoT-SORT~\citep{aharon2022BoTSORTRobust} & \textit{\textcolor{blue}{64.6}} & NA & NA & \textit{\textcolor{blue}{80.6}} & \textit{\textcolor{blue}{79.5}} \\
SparseTrack~\citep{liu2023SparseTrackMultiObject} & \textbf{\textcolor{red}{65.1}} & \textbf{\textcolor{red}{65.3}} & \textbf{\textcolor{red}{65.1}} & \textbf{\textcolor{red}{81.0}} & \textbf{\textcolor{red}{80.1}} \\
UCMCTrack~\citep{yi2024UCMCTrackMultiObject} & 64.3 & NA & \textit{\textcolor{blue}{64.6}} & 79.0 & 79.0 \\
\midrule
\multicolumn{6}{l}{\tiny{METHODS NOT USING ANY MOTION COMPENSATION}} \\
SORT~\citep{bewley2016Simpleonline}             & 63.0 & 64.2 & 62.2 & 80.1 & \textit{\textcolor{blue}{78.2}} \\
ByteTrack~\citep{zhang2021ByteTrackMultiObject}        & 63.1 & \textit{\textcolor{blue}{64.5}} & 62.0 & \textit{\textcolor{blue}{80.3}} & 77.3 \\
OC-SORT~\citep{caoObservationCentricSORTRethinking2022}          & \textit{\textcolor{blue}{63.2}} & 63.2 & \textit{\textcolor{blue}{63.2}} & 78.0 & 77.5 \\
MotionTrack~\citep{xiao2023MotionTrackLearning} & 60.9 & NA & 59.4 & 76.5 & 73.5 \\
C-BIoU~\citep{yang2023HardTrack}           & \textbf{\textcolor{red}{64.1}} & \textbf{\textcolor{red}{64.8}} & \textbf{\textcolor{red}{63.7}} & \textbf{\textcolor{red}{81.1}} & \textbf{\textcolor{red}{79.7}} \\
C-TWiX (ours)           & 63.1 & 64.1 & 62.5 & 78.1 & 76.3 \\

\bottomrule
\end{tabular}
\end{center}
\end{table}

\begin{table*}[ht]
\begin{center}
\caption{Performance on the test set of KITTIMOT. Only trackers using the detections from Permatrack and using only coordinates are shown. \textbf{\textcolor{red}{Bold red}} and \textit{\textcolor{blue}{italic blue}} indicate respectively the first and second best methods within each category.}
\label{tab:results_test_kittimot}
\begin{tabular}{lp{0.8cm}p{0.7cm}p{1cm}|p{0.8cm}p{0.7cm}p{1cm}}
 & \multicolumn{3}{c}{car} & \multicolumn{3}{c}{pedestrian} \\
Method & \footnotesize{HOTA} & \footnotesize{AssA} & \footnotesize{MOTA} & \footnotesize{HOTA} & \footnotesize{AssA} & \footnotesize{MOTA}\\
\toprule
\multicolumn{4}{l}{\tiny{METHODS USING MOTION COMPENSATION}} & &  &\\
\footnotesize{Permatrack~\citep{tokmakov2021LearningTrack}} & \textbf{\textcolor{red}{77.4}} & \textbf{\textcolor{red}{77.7}} & \textbf{\textcolor{red}{90.9}} & \textit{\textcolor{blue}{47.4}} & \textit{\textcolor{blue}{43.7}} & \textit{\textcolor{blue}{65.1}} \\
\footnotesize{UCMCTrack~\citep{yi2024UCMCTrackMultiObject}} & \textit{\textcolor{blue}{77.1}} & \textit{\textcolor{blue}{77.2}} & \textit{\textcolor{blue}{90.4}} & \textbf{\textcolor{red}{55.2}} & \textbf{\textcolor{red}{58.0}} & \textbf{\textcolor{red}{67.4}} \\
\midrule
\multicolumn{4}{l}{\tiny{METHODS NOT USING ANY MOTION COMPENSATION}} & & &\\
\footnotesize{OC-SORT~\citep{caoObservationCentricSORTRethinking2022}}  & \textit{\textcolor{blue}{74.6}} & \textit{\textcolor{blue}{74.5}} & \textit{\textcolor{blue}{87.8}} & \textbf{\textcolor{red}{53.0}} & \textbf{\textcolor{red}{57.8}} & \textit{\textcolor{blue}{62.0}} \\
\footnotesize{C-TWiX (ours)}     & \textbf{\textcolor{red}{77.6}} & \textbf{\textcolor{red}{78.8}} & \textbf{\textcolor{red}{89.7}} & \textit{\textcolor{blue}{52.4}} & \textit{\textcolor{blue}{54.4}} & \textbf{\textcolor{red}{65.0}} \\

\bottomrule
\end{tabular}
\end{center}
\end{table*}

On MOT17, the results of our tracker C-TWiX are on par with other trackers that do not rely on a camera motion compensation technique. Since this dataset contains many sequences with a moving camera, estimating the motion of the camera particularly helps the tracker. 

Moreover, on KITTIMOT, where the objects of interest are cars and pedestrians, our tracker C-TWiX outperforms other trackers on cars, even those using motion compensation methods, and gets competitive results on pedestrians. Permatrack reaches excellent results on cars to the detriment of pedestrians, while OC-SORT shows the opposite trade-off. With C-TWiX, the average HOTA score increases by 0.6 point. Even if the framerate on this dataset is low, TWiX manages to correctly associate pedestrians, which is the hardest class. Indeed, they have a vertically elongated shape but move mainly on the horizontal axis that makes the overlap of bounding boxes between adjacent frames very low.

Finally, our tracker is fast running at 320 Hz on KITTIMOT, 300 Hz on DanceTrack and 50 Hz on MOT17 on a single GeForce RTX 2060 (without considering the detection part).

\subsection{Ablation study}

In the following, we evaluate the effect of our choices in the design of the architecture of the TWiX module, measure the performance of the tracker C-TWiX on ground-truth detections, and visualize how the affinity between two bounding boxes evolves with regard to their relative position.

\subsubsection{Oracle detections}

Similarly to previous studies~\citep{sun2021DanceTrackMultiObject, yang2023HardTrack}, we conducted an experiment by replacing the detections by the ground-truth annotations, to evaluate only the association component of the tracking. Table~\ref{tab:results_oracleDet} indicates that our C-TWiX tracker surpasses all other methods based on positions or motion on the DanceTrack dataset. The HOTA score is improved by 0.4 point compared to the previous best method and the DetA by 1 point, thanks to a better association. 

\begin{table}[ht]
\begin{center}
\caption{Performance on the validation set of DanceTrack using oracle detections. \textbf{\textcolor{red}{Bold red}} and \textit{\textcolor{blue}{italic blue}} indicate respectively the first and second best methods.}
\label{tab:results_oracleDet}
\begin{tabular}{lccccc}
Loss function & HOTA & DetA & AssA & MOTA & IDF1 \\
\midrule
IoU~\citep{sun2021DanceTrackMultiObject}           & 72.8          & \textbf{\textcolor{red}{98.9}}          & 53.6          & 98.7          & 63.5 \\
IoU + Motion~\citep{sun2021DanceTrackMultiObject}  & 69.4          & 87.9          & 54.8          & 99.4          & 71.3 \\
SORT~\citep{bewley2016Simpleonline}                & 67.6          & 86.6          & 52.8          & 98.1          & 69.6\\
OC-SORT~\citep{caoObservationCentricSORTRethinking2022}       & 79.1          & 97.7          & \textit{\textcolor{blue}{64.0}}          & \textit{\textcolor{blue}{99.6}}          & 76.1\\
C-BIoU~\citep{yang2023HardTrack}        & \textit{\textcolor{blue}{81.7}}         & 97.6          & \textbf{\textcolor{red}{68.4}} & 99.3          & \textbf{\textcolor{red}{80.5}}\\
C-TWiX (ours)        & \textbf{\textcolor{red}{82.1}}& \textit{\textcolor{blue}{98.6}} & \textbf{\textcolor{red}{68.4}} & \textbf{\textcolor{red}{99.7}} & \textit{\textcolor{blue}{78.1}}\\
\bottomrule
\end{tabular}
\end{center}
\end{table}

\subsubsection{Inter-Pair Transformer Encoder}

\begin{figure}[h] \centering
    \includegraphics[trim=0 0 0 0, width=13cm, clip]{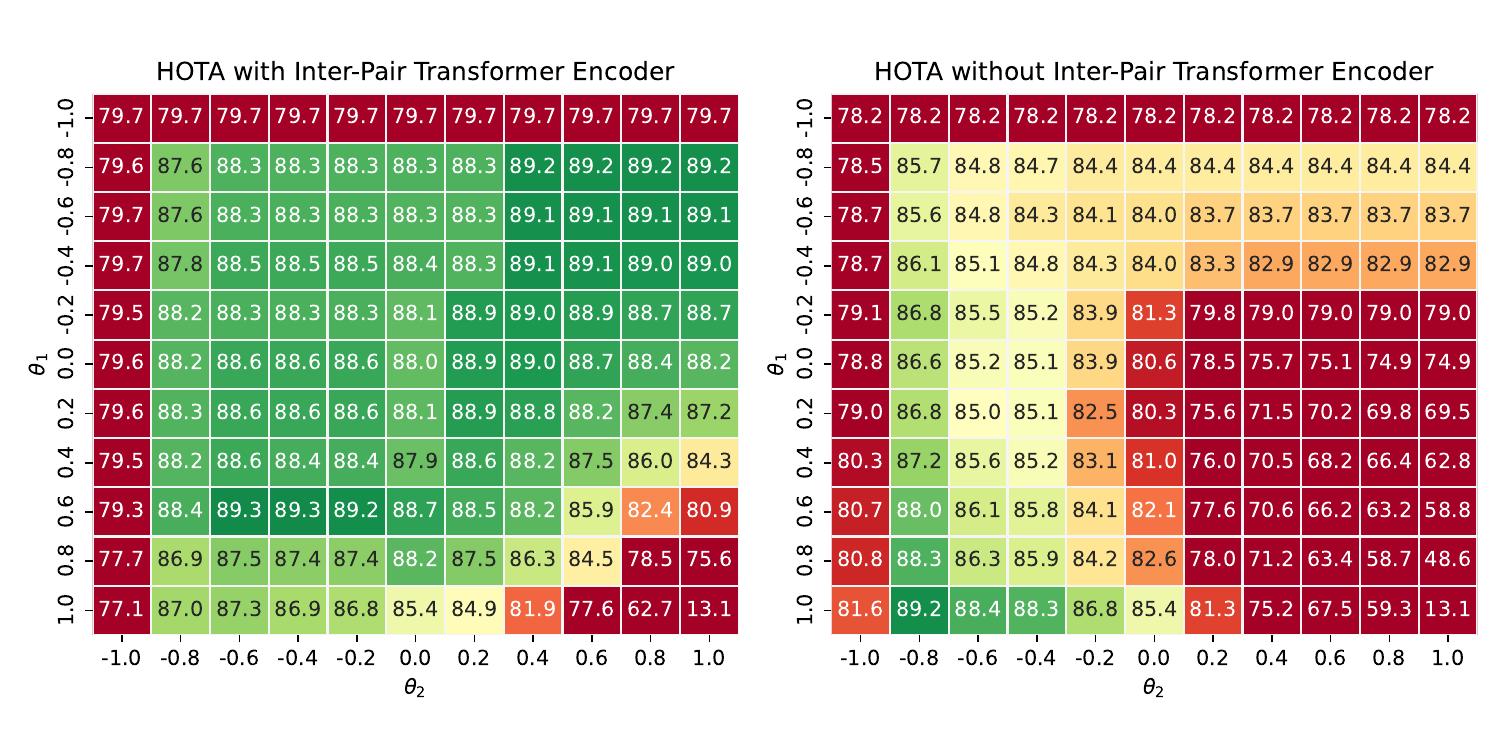}
    \caption{Comparison of HOTA on the validation set of KITTIMOT-car at different level of matching regarding the presence of the Inter-Pair Transformer Encoder (left) or not (right).}
    \label{fig:interpair}
\end{figure}

In the architecture of the TWiX module, the Inter-Pair Transformer Encoder aims to enhance the representation of the pair embeddings. To verify this assertion, we trained the TWiX modules with and without this second Transformer Encoder. Then, we evaluate the HOTA score at different thresholds $(\theta_1, \theta_2)$ on the validation set of KITTIMOT. Figure~\ref{fig:interpair} shows the heatmap of the HOTA score at different thresholds. Specifically, when $\theta_1 = 1$ (resp. $\theta_2 = 1$), only the second (resp. first) matching step is on, rejecting all possible matching at the first (resp. second) association step. Without the Inter-Pair Transformer Encoder, the HOTA score of cars drops by 6 points in the case of a pipeline with only the first matching step ($\theta_2 = 1$). This means that the quality of the first TWiX module, operating on adjacent tracklets, becomes poor at discriminating positive pairs from negative one. In that situation, since coordinates of bounding boxes are close, the absence of a layer interacting with all pairs degrades the quality of the module. 

However, in the case where only the second matching step is on ($\theta_1 = 1$), the absence of the Inter-Pair Transformer Encoder does not provoke a drop in HOTA score. Since the second TWiX module was trained on harder batches ($t_G \neq 0$) and is deeper (4 layers instead of 1), its discriminative power is higher than the first one, at the cost of slower calculation time. The Inter-Pair Transformer Encoder is therefore less essential in that case. 

Nevertheless, adding the Inter-Pair Transformer Encoder increases the area in the $(\theta_1, \theta_2)$ space with HOTA over 89. This results in a green zone which covers a larger part, whereas without this encoder, such a HOTA score is reachable only when $\theta_1$ equals 1. In conclusion, the Inter-Pair Transformer Encoder makes the tracker more robust with respect to the selection of the hyper-parameters $(\theta_1$ and $\theta_2)$, and also makes the tracking faster since the first TWiX module can rely on a single-layer Transformer.


\subsubsection{Loss Function}

The TWiX modules are trained with a bidirectional contrastive loss, as referred in Equation~\ref{eq:loss_bidirectional}. The use of such a function is motivated to force a positive pair to have a higher affinity than any other negative pair within the same row or same column in the affinity matrix. In order to validate the choice of the bidirectional contrastive loss, we conducted an experiment with six other loss functions, keeping all the other hyper-parameters identical, except $\theta_1$ and $\theta_2$.

\begin{figure*}[ht] \centering
    \includegraphics[trim=0 0 0 0, width=14cm, clip]{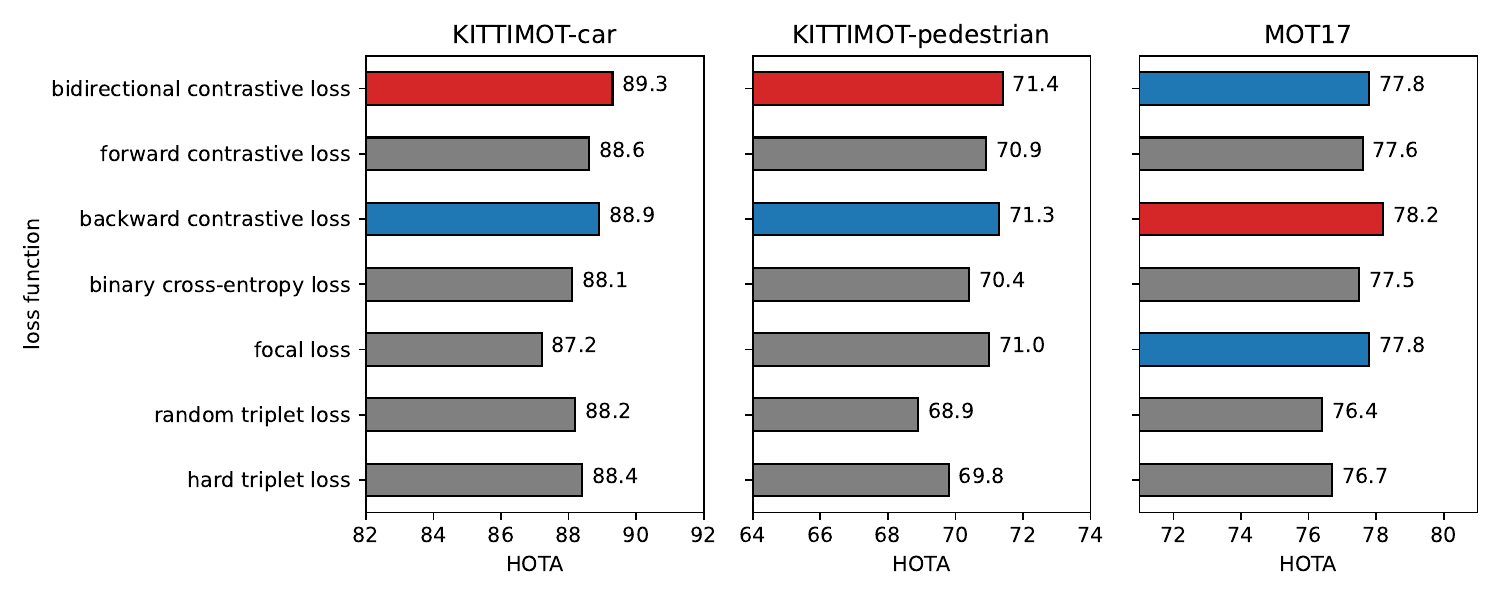}
    \caption{HOTA scores on KITTIMOT and MOT17 validation sets with regard to the loss function. \textcolor{red}{Red} and \textcolor{blue}{blue} indicate respectively the first and second best methods.}
    \label{fig:twix_loss}
\end{figure*}

Experiments are conducted on the KITTIMOT and MOT17 datasets and reported in Figure~\ref{fig:twix_loss}. The parameters $\theta_1$ and $\theta_2$ are selected with a grid search for each dataset and object class.

First, since the affinity matrix returns values between -1 and 1 for each pairs of tracklets, we use the binary cross-entropy after linearly scaling the values to the range from 0 to 1. Despite the unbalanced ratio between the positive and negative pairs (1:20), using the binary cross-entropy loss reaches decent results with a drop of respectively 1.2, 1.0 and 0.3 points in HOTA score on respectively KITTIMOT cars, pedestrians and on MOT17 compared to the use of our bidirectional contrastive loss.

The focal loss was introduced to overcome the unbalanced data ratio~\citep{lin2017FocalLoss}. This improves the HOTA on pedestrians on KITTIMOT and MOT17 by 0.6 and 0.3, but decreases it on cars by 0.9 point compared to the use of the binary cross-entropy loss. However, these results are all lower than those obtained with our contrastive loss.

Both binary cross-entropy and focal losses compute the loss for each pair regardless of other pairs in the affinity matrix. Triplet loss aims at maximizing a positive pair affinity compared to a single negative pair, over a certain margin. For each positive pair in the matrix, we randomly select a negative pair on the same row or same column: this is the random triplet loss. To take into account harder negatives, as a critical component of contrastive learning~\citep{appalaraju2020GoodPractices}, we select the negative pair on the same row or column which has the highest affinity: this is the hard triplet loss. We notice a drop in performance with both losses compared to our bidirectional contrastive loss. Using the hard negative instead of a random negative increases the HOTA between 0.2 and 0.9: making the learning harder by forcing the TWiX module to discriminate harder cases improves the performance of the model. But considering all negatives, like in our contrastive loss, improves the quality of our association module.

And finally, since the bidirectional contrastive loss is the sum of the forward and the backward contrastive losses, we compared it with its two components. Except on MOT17 for the backward version, our loss function beats the forward and the backward versions. This validates that a signal containing both negative from the same row (forward version) and from the same column (backward version) improves the learning by providing harder cases, since the model needs to discriminate a positive pair from all other negative pairs.

\subsubsection{Visualization of TWiX Self-Affinity Maps}

Contrarily to other IoU-based methods, TWiX takes into account the whole neighborhood of a bounding box to compute its affinities. That is why, in practice, it is hard to visualize the affinity matrix for TWiX between two sets of detections. Moreover, TWiX can leverage the temporal aspect, what other methods cannot do. Therefore, we propose, in a matter a simplicity, to measure the self-affinity of a bounding box: given a box, we compute the affinity between itself and a translated version. The translation is added on the box vertically and horizontally. The obtained map indicates the locations where the affinity is the highest. 

\begin{figure}[h] \centering
    \includegraphics[trim=0 0 0 0, width=14cm, clip]{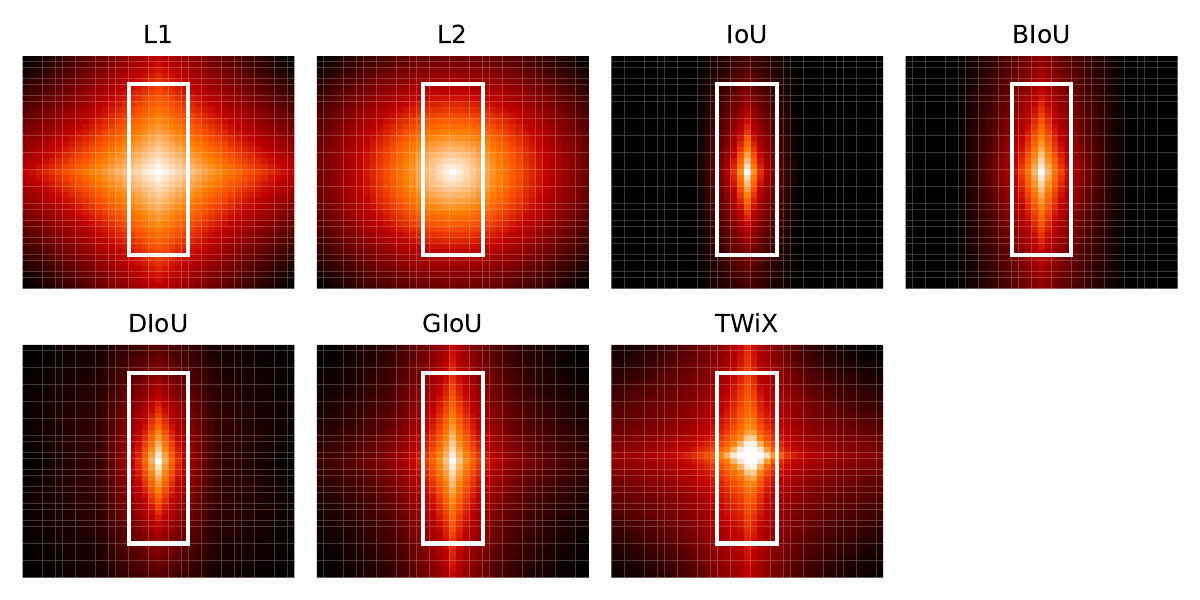}
    \caption{Self-affinity maps of several model-based methods and TWiX. The affinity between the white box of reference and its translated version is indicating by the color at its translated center position. The whiter, the higher the affinity is.}
    \label{fig:twix_vs_others}
\end{figure}

Figure~\ref{fig:twix_vs_others} illustrates these maps for six model-based methods and TWiX, which is data-based. First, for the distance $L_1$ and $L_2$, symmetrical diamond-shaped and circular maps are obtained respectively, ignoring the shape of the original box. For IoU and the buffered version BIoU~\citep{yang2023HardTrack}, we notice that the map extends in the predominant direction of the box, here the vertical axis. This behaviour is not desired because the motion of an object is not related to its shape. Moreover, the map reaches zero at an intermediate distance, where the overlap between the boxes is empty. Methods such as DIoU~\citep{zheng2020DistanceIoULoss} and GIoU~\citep{rezatofighi2019GeneralizedIntersection} solve this last issue, but a preferential direction is still present based on the shape of the object. Finally, for TWiX, not only the map takes the shape of a diffraction spike with two preferential directions but also, it does not reach zero at an intermediate distance. This behaviour is desired because it means that TWiX learns that objects mainly move vertically or horizontally, the actual direction of motion. 

\begin{figure}[h] \centering
    \includegraphics[trim=0 0 0 0, width=10cm, clip]{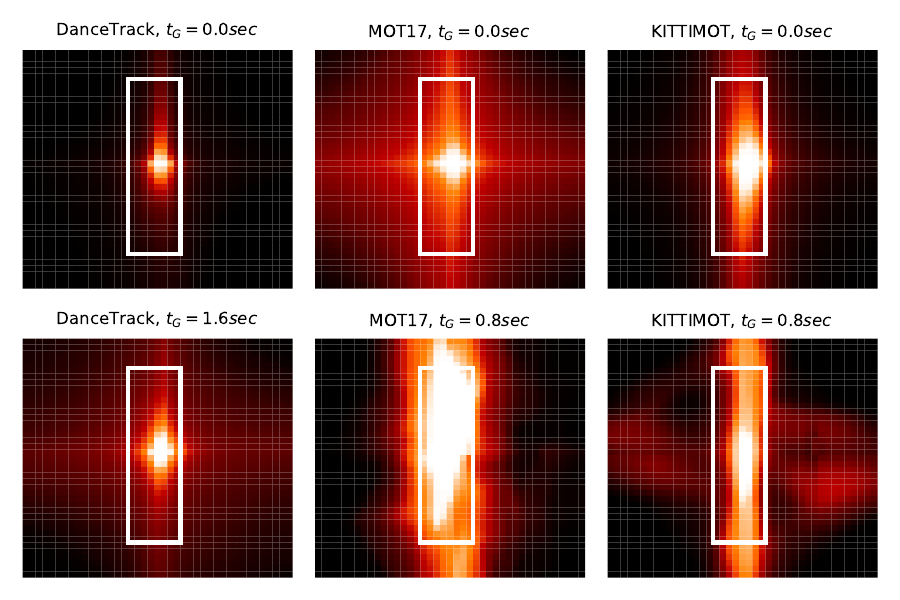}
    \caption{Self-affinity maps of TWiX on different datasets and with different maximal temporal gaps $t_G$}
    \label{fig:twix_vs_datasets}
\end{figure}

Moreover, TWiX adapts to the dataset (framerate, object motion, camera motion, object size, etc.) and to the maximal temporal gap allowed. Figure~\ref{fig:twix_vs_datasets} illustrates this. In the case of a dataset with high framerate, the shape of the spike will be smaller indicating small motion, such as in DanceTrack ($fps=30$), contrarily to KITTIMOT ($fps=10$) where a bigger spike is observed. And allowing a higher temporal gap $t_G$ enlarges the map, expanding the search area for a possible match.

\section{Conclusion}

In this work, we proposed a contrastive framework to learn representations on pairs of tracklets for MOT. Contrastive learning has shown promising results to learn representations in textual data, images and online MOT. To the best of our knowledge, this is the first work exploiting contrastive learning on the association step of a tracker solely based on coordinates. Our framework creates batches of tracklets that are later encoded with two Transformer Encoders. The first one extracts representation for a pair of tracklets and the second Encoder learns to enhance these representations by paying attention to every pair of tracklets. Experiments on multiple datasets show that our tracker C-TWiX outperforms previous methods on DanceTrack and KITTIMOT and is on par on MOT17. Contrarily to other IoU-based approaches, our module TWiX is able to learn motion from tracklets adjusting the search area for each object.

Even if our module TWiX requires to create all pairs of tracklets, resulting in a bi-quadratical computation during the attention mechanism, our tracker C-TWiX is able to track objects in real time.

\section*{Acknowledgments}

We acknowledge the support of the Natural Sciences and Engineering Research Council of Canada (NSERC) [funding reference number RGPIN-2020-04633]. 



\bibliographystyle{elsarticle-harv}
\bibliography{references}




\end{document}